\documentclass{article}
\usepackage{ijcai18}
\usepackage{algorithm}
\usepackage[noend]{algpseudocode}
\usepackage{times}
\usepackage{soul}
\usepackage{url}
\usepackage[hidelinks]{hyperref}
\usepackage[utf8]{inputenc}
\usepackage[small]{caption}
\usepackage{amsmath}
\usepackage{subfig}
\usepackage{graphicx}
\usepackage{booktabs,array,dcolumn}
\newcolumntype{d}{D{.}{.}{2.3}}
\newcolumntype{C}{>{\centering}p}
\title{In Hindsight: A Smooth Reward for Steady Exploration}

\author{
Hadi Samer Jomaa$^1$,
Josif Grabocka$^1$,
Lars Schmidt-Thieme$^1$
\\ 
$^1$ Information Systems and Machine Learning Lab, University of Hildesheim, Germany\\
\{hsjomaa,josif,schmidt-thieme\}@ismll.uni-hildesheim.com
}
\begin{document}

\maketitle

\begin{abstract}
In classical Q-learning, the objective is to maximize the sum of discounted rewards through iteratively using the Bellman equation as an update, in an attempt to estimate the action value function of the optimal policy. Conventionally, the loss function is defined as the temporal difference between the action value and the expected (discounted) reward, however it focuses solely on the future, leading to overestimation errors. We extend the well-established Q-learning techniques by introducing the hindsight factor, an additional loss term that takes into account how the model progresses, by integrating the historic temporal difference as part of the reward. The effect of this modification is examined in a deterministic continuous-state space function estimation problem, where the overestimation phenomenon is significantly reduced and results in improved stability. The underlying effect of the hindsight factor is modeled as an adaptive learning rate, which unlike existing adaptive optimizers, takes into account the previously estimated action value. 
The proposed method outperfoms variations of Q-learning, with an overall higher average reward and lower action values, which supports the deterministic evaluation, and proves that the hindsight factor contributes to lower overestimation errors. The mean average score of 100 episodes obtained after training for 10 million frames shows that the hindsight factor outperforms deep Q-networks, double deep Q-networks and dueling networks for a variety of ATARI games.
\end{abstract}

\section{Introduction}

Reinforcement learning (RL) has gained considerable attention over the past five years. In this field of research, an agent attempts to learn a behavior through trial-and-error interactions with a dynamic environment, in order to maximize an allocated reward, immediate or delayed. Broadly speaking, an agent selects an optimal policy either by using an optimal value function, or by manipulating the policy directly. Thanks to the rich representational power of neural networks, the high-dimensional input obtained from real-world problems can be reduced to a set of latent representations with no need for hand-engineered features. Recently, a variant of Q-learning based on convolutional neural networks demonstrated remarkable results on a majority of games within the Arcade Learning Environment, by reformulating the RL objective as a sequential supervised learning task \cite{mnih2015human}. One of the contributing factors to this approach is the presence of an experience replay memory, which stores the transitions at every step. This leads to the temporal de-correlation between experiences, and hence upholds the i.i.d assumption which allows stochastic gradient-based learning \cite{lin1992self}.  Experience replay reduces the amount of episodes required for training \cite{schaul2015prioritized} even though some transitions might not be immediately useful \cite{schmidhuber1991curious}.
A well-defined optimal value function also plays a major role in RL tasks. At its core, an optimal value function is basically an approximation of the predicted reward given a state-action pair. Higher rewards are hence achieved by navigating the environment, acting greedily with respect to the value function. \\
In this paper, we reshape the reward as a weighted average between the expected discounted reward, for a sampled state-action pair, and its previously selected action-value. In one-step Q-learning, the loss, $L^Q$, is calculated based on the temporal difference (TD) between the discounted reward and the estimated action-value at any given state, Equation \ref{td}:
\begin{equation}
\label{td}
L^Q(\theta_i) = \bigl(r_j + \gamma\max_{a'}\mathcal{Q}_\text{target}(s_{j+1},a';\theta^-) - \mathcal{Q}(s_j,a_j;\theta_i)\bigr)^2
\end{equation}
where $s_j$, $a_j$, and $r_j$ represent respectively the state, action, and reward at sampled iteration $j$, with $\gamma$ as the discount factor, and $\theta_i$ as the network parameters at current iteration $i$, such that $j<i$, and $\theta^-$ as the target network parameters.\\
One drawback of this equation is that it only focuses on the forward temporal difference, i.e. between the action-value at the current sampled state, $s_j$ and the next state $s_{j+1}$. It doesn't however address how the decision process has evolved between updates by ignoring action values at previous iterations. We leverage this information under the assumption that it helps lower the variance of action-values throughout the learning process, while concurrently maintaining high rewards, and ultimately learn a better agent.\\
To that end, we introduce an additional term to the standard TD error, referred to as the hindsight factor, $L^H(\theta_i)$, that represents the difference between the action-value of the sampled state at the current iteration and its previously selected (stored) action-value, obtained with network parameters $\theta_{j}$:
\begin{equation}
L^H(\theta_i) = \bigl(\mathcal{Q}(s_j,a_j;\theta_i) - \mathcal{Q}(s_j,a_j;\theta_j)\bigr)^2
\end{equation}
The key observation is that the hindsight factor acts as regularizer on the Q-network, unlike conventional regularization techniques that force restrictions on the network parameters directly. The hindsight factor can also be considered as an adaptive learning step controller that penalizes large deviations from previous models by incoporating the momentum of change in action values across updates. Inherently, if the hindsight factor increases, this means that the model parameters have significantly changed, leading to higher (or lower) action-values. The introduction of the hindsight factor restructures the reward, as a weighted average between what is expected, and what was estimated, ensuring that model updates are carried out cautiously.%, while more grounded Q-values.
\\
We summarize the contributions of the paper as follows:
\begin{itemize}
\item A novel extension to the optimal value function technique, which leads to an overall improved performance
\item Deterministic evaluation on a continuous state-space that shows how the hindsight factor reduces the bias in function approximation 
\item Experiments on ATARI games that highlight the effect of the hindsight factor
\item Comparative analysis that demonstrates the effect of adding the hindsight factor to multiple variations of deep Q-networks
\end{itemize}

\section{Background}
A standard reinforcement learning setup \cite{sutton2018reinforcement}, consists of an agent interacting with an environment $\mathcal{E}$ at discrete timesteps. This formulation is based on a Markov Decision Process (MDP) represented by $\langle S, A, R, T\rangle$. At any given timestep $j$, the agent receives a state $s_j \in S$, upon which it selects an action $a_j \in A$, and a scalar reward $r_j \in R$ is observed. The transition function $T : S\times A \rightarrow S$ generates a new state $s_{j+1}$. The agents behaviour is governed by a policy, $\pi: S \rightarrow A$, which computes the true state-action value, as
\begin{equation}
Q_{\pi}(s,a) = E_{\pi}\left[\Sigma_{t=0} \gamma^tr^t  | S_0 = s, A_0 = a\right], 
\end{equation}
where $\gamma \in [0,1]$ represents the discount factor balancing between immediate and future rewards.\\
To solve this sequential decision problem, the optimal policy selects the action that maximizes the discounted cumulative reward, $\pi_{\ast}(s) \in \text{arg max}_a Q_{\ast}(s,a)$, where $Q_{\ast}(s,a) = \max_{\pi} Q_{\pi}(s,a)$ denotes the optimal action value.\\
One of the most prominent value-based methods for solving reinforcement learning problems is Q-learning \cite{watkins1992q}, which directly estimates the optimal value function and obeys the fundamental identity, known as the Bellman equation \cite{bellman1957functional}
\begin{equation}
Q_{\ast}(s,a) = E\left[r_j + \gamma \max_{a'}Q_{\ast}(s_{j+1},a') | S_0 = s, A_0 = a\right]
\end{equation}
As the number of states increases, learning all action values per state separately becomes computationally taxing, which is why the value function is approximated via a paramatrized network, resulting in $Q(s,a) = Q(s,a;\theta)$. 

\subsection{Deep Q-networks}
Deep Q-network (DQN) is a model-free algorithm presented by \cite{mnih2015human}, which learns the Q-function in a supervised fashion. The objective is to minimize the loss function,
\begin{equation}
L^Q_i(\theta_i) = E_{s,a\sim p(.)}[\left(y_j - Q(s,a;\theta_i)\right)^2],
\end{equation}
with $p(.)$ as the probability distribution over the action space, and the target $y_j$ as the expected discounted reward, Equation \ref{y_i}:
\begin{equation}
\label{y_i}
y_j=E_{s' \sim \mathcal{E}}\left[r_j + \gamma \max_{a'}Q_{target}(s',a;\theta^-)\right]. 
\end{equation}
In this approach, a target network, $Q_{target}(s',a',\theta^{-})$ shares the same architecture as the online network $Q(s,a;\theta_i)$, however it is only updated after a fixed number of iterations which increases the stability of the algorithm. The correlation between sequential observations is reduced by uniformly sampling transitions of the form $(s_j,s_{j+1},r_j,a_j)$ for the off-policy from a replay buffer \cite{lin1993reinforcement}.
\subsection{Double Deep Q-networks}
Double DQN (DDQN) \cite{van2016deep} is a variation of DQN, where the action selected and its corresponding value are obtained from two separate networks. In other words, based on the sampled state $s_{j}$, the action is selected based on the greedy policy of the network, i.e. $a_{j+1} = \text{arg}\max_{a'} Q(s_{j+1},a',\theta)$, whereas the reward is calculated based on the action-value of the selected from the target network, resulting in the following target, 
\begin{equation}
y_j = E_{s' \sim \mathcal{E}}\left[r_j + \gamma Q_\text{target}(s',a_{j+1};\theta^-)\right]. 
\end{equation}
This comes as a solution to the overoptimistic value estimates, which result from using the same network to select and evaluate the action. 
\subsection{Dueling Network Architectures}
Sharing similar lower level representation, the dueling architecture (DUEL) \cite{wang2015dueling} extends DQN by explicitly separating the representation of the state values from the state-dependent action values. This allows the network to understand which states are valuable agnostic to the action selected, which is particularly important for setups where the action does not have a major effect on the environment. The action value is estimated as a function of two modules, the action advantage module $Adv$, and the state-value $V$ as expressed in Equation \ref{duelexp}:
\begin{equation}
\label{duelexp}
\begin{split}
Q(s,a;\theta,\alpha,\beta) = & \left(Adv(s,a;\theta,\alpha) - \frac{1}{|Adv|}\Sigma_{a'}Adv(s,a';\theta,\alpha)\right)\\
&+V(s;\theta,\beta)
\end{split}
\end{equation}
with $\theta$ as the shared parameters, $\beta$ as the parameters of the state-value module, and $\alpha$ as the parameters of the action value module.
%The relative rank of the action-value is not affected when subtracting the mean, and hence the greedy policy is preserved, with the optimal action  $a_{\ast} = \text{arg max}_{a' \in A} Q(s,a';\theta,\alpha,\beta) = \text{arg max}_{a' \in A} A(s,a';\theta,\alpha)$.
\\
Other variations that have been proposed on Q-networks include selecting an action based on the average of it value over across previous networks \cite{anschel2017averaged} which preserves the original loss function, and $l_2$-regularization \cite{farebrother2018generalization} to avoid overfitting on the training environment.\\
The proposed modification to the existing Q-learning networks shares the same input-output interface, however, it reformulates the existing loss function $L^Q(\theta_i)$, where the objective is to minimize the difference between the action-value and the future discounted reward, by introducing the hindsight factor $L^H(\theta_i)$ that adaptively manages reward expectation.
\begin{equation}
\label{L1}
%{L^H}_i(\theta_i) =  E_{s,a\sim p(.)}\left[\bigl(\overbrace{\mathcal{Q}(s_j,a_j,\theta_j) - \mathcal{Q}(s_j,a_j;\theta_i)}^\text{hindsight factor}\bigr)^2\right]
{L^H}(\theta_i) =  \bigl(\overbrace{\mathcal{Q}(s_j,a_j,\theta_j) - \mathcal{Q}(s_j,a_j;\theta_i)}^\text{hindsight factor}\bigr)^2
\end{equation}\\
\begin{figure*}[h]
\subfloat[True value and an estimate]{%
  \includegraphics[width=.7\columnwidth]{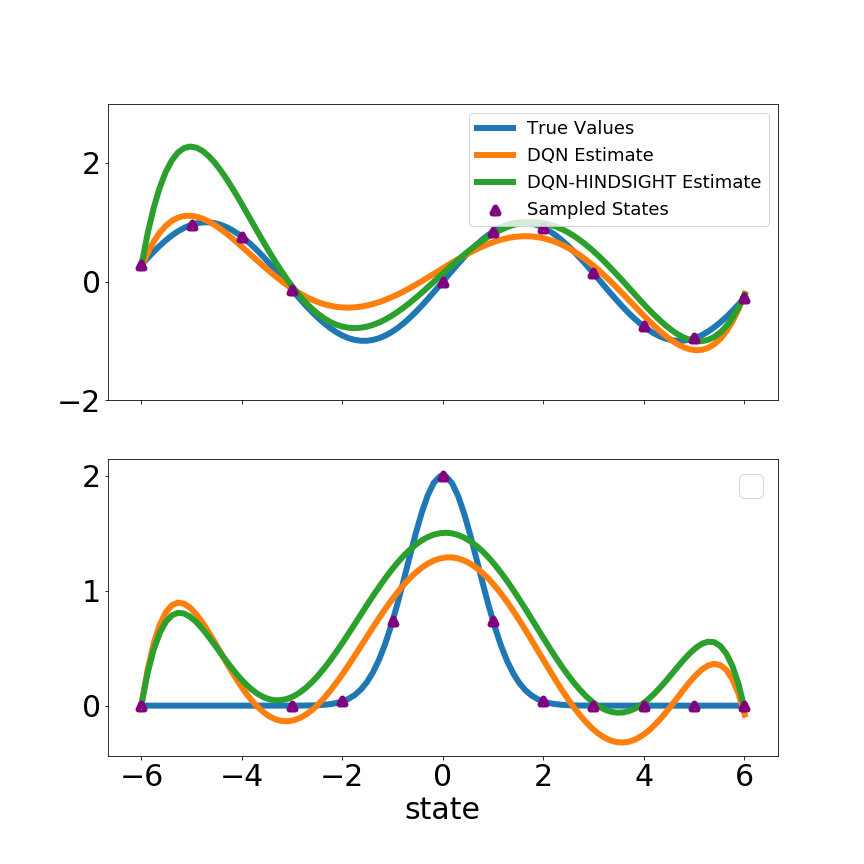}%
}
\subfloat[Bias in DQN as a function of state]{%
  \includegraphics[width=.7\columnwidth]{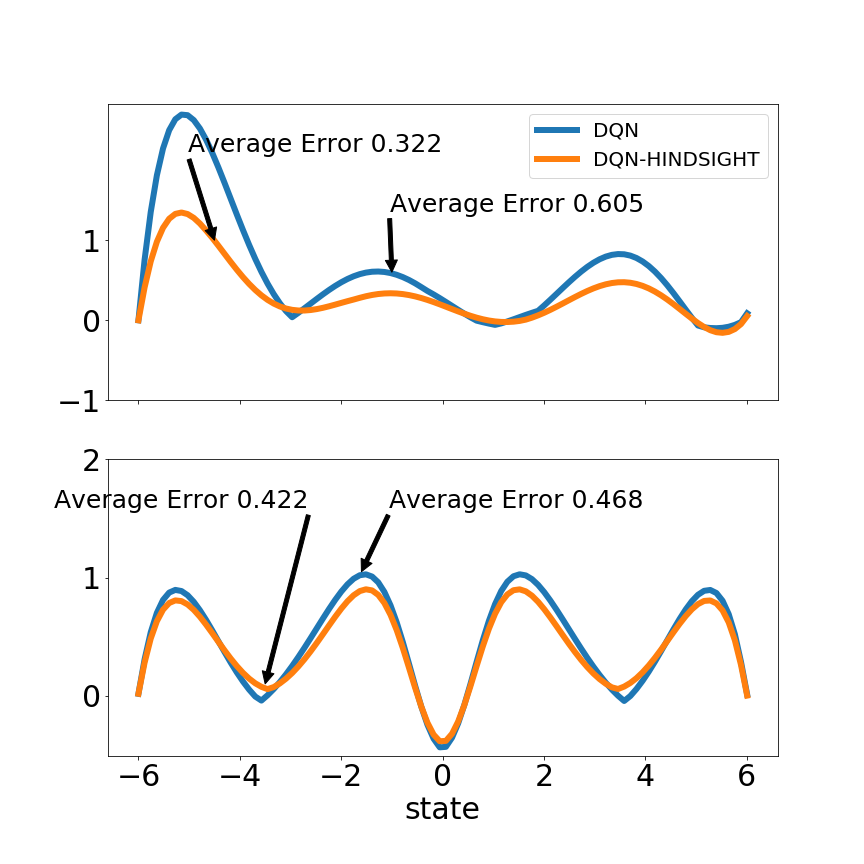}%
}
\subfloat[Bias in DDQN as a function of state]{%
  \includegraphics[width=.7\columnwidth]{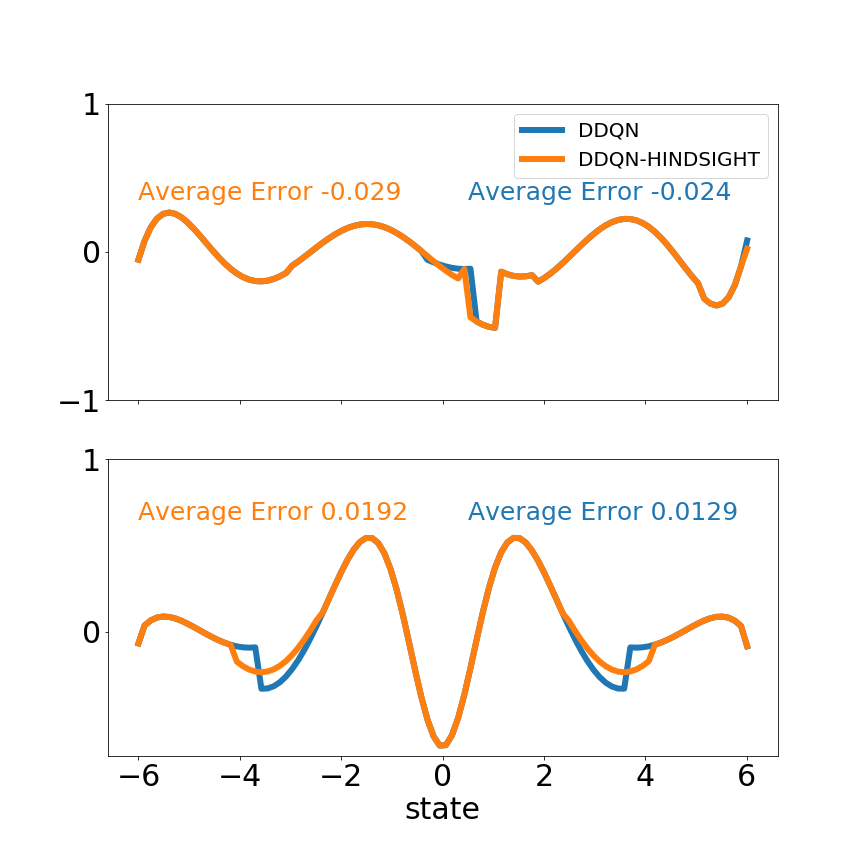}%
}
\caption{Illustration of Overestimations}
\label{overest}
\end{figure*}
\section{In Hindsight}
Hindsight is an extension of the conventional value iteration techniques in reinforcement learning that considers the previous performance of the network when calculating the reward. In this supervised formulation, the target is calculated based on the forward directional temporal difference, i.e. between the estimated action-value and the expected discounted reward (future), Equation \ref{td},
%\begin{multline}
%\label{L0}
%L^{B}_i(\theta_i) = E_{s,a\sim p(.)}[(\underbrace{R_j + \gamma\max_{a'}\mathcal{Q}_\text{target}(s_{j+1},a';\theta^-)}_\text{y} - 
%\\ \underbrace{\mathcal{Q}(s_j,a_j;\theta_i)}_{\hat{y}})^2 ]
%\end{multline}    
%The proposed method outperfoms the widely adopted variations of Deep Q-Networks, and is tested across more than 30 ATARI games with similar hyperparameters. We notice that not only the overall average reward is higher than the corresponding techniques, but also the action-values are smaller and monotonically increasing as function of frames.     
dropping any use of previous action-state value (past). To counter that effect, we introduce the hindsight factor, Equation \ref{L1}, to balance the current action-value estimate, and prevent overestimation. \cite{thrun1993issues}
Intuitively, this term represents the confidence of the agent in previous actions. More specifically, if the estimated action-value at current iteration $i$, is much higher than the previously estimated action-value at iteration $j$, $q_j = \max \mathcal{Q}(s_j,a_j;\theta_j)$, given the same state representation $s_j$, then \textit{in hindsight} action $a_j$ was not optimal in the global sense, even though it was selected based on greedy policy as $a_j = \text{arg max}_a Q(s_j,\theta_j)$. If on the other hand, the historical temporal difference is small, then the network is converging in the optimal direction, as given the same state and the same action, the corresponding action value is equally high.\\
The total loss would simply be a weighted sum between the forward temporal difference, $L^D$ and the backward temporal difference, $L^H$. 
If we expand the loss and factorize the components, we end up with a new loss representation, Equation \ref{Lnew},
\begin{equation}
\label{Lnew}
\begin{split}
L(\theta_i) = & \bigl(\overbrace{r_j + \gamma\max_{a'}\mathcal{Q}_\text{target}(s_{j+1},a';\theta^-)}^{\hat{y}} - \mathcal{Q}(s_j,a_j;\theta_i)\bigr)^2 + \\ & \delta\bigl(\overbrace{\mathcal{Q}(s_j,a_j,\theta_j)}^{\bar{y}} - \mathcal{Q}(s_j,a_j;\theta_i)\bigr)^2
%L_{i} = E_{s,a\sim p(.)}\left[\bigl(\overbrace{\frac{y+\delta q_j}{1+\delta}}^{R_\text{new}} -  \mathcal{Q}(s_j,a_j;\theta_i)\bigr)^2\right], 
\end{split}
\end{equation}
with $\delta$ as the hindsight coefficient.  The hindsight factor inherently restructures the target reward as a smooth trade-off between the expected and the previously estimated reward. This is derived by first expanding the loss, Equation \ref{expand},
\begin{equation}
\label{expand}
\begin{split}
L(\theta_i) = &\hat{y}^2 -2\hat{y}\mathcal{Q}(s_j,a_j;\theta_i) + \mathcal{Q}(s_j,a_j;\theta_i)^2 + \\ &
\delta\bar{y}^2 -2\delta\bar{y}\mathcal{Q}(s_j,a_j;\theta_i) + \delta\mathcal{Q}(s_j,a_j;\theta_i)^2
\end{split}
\end{equation}
since $\hat{y}$ and $\bar{y}$ are simply constants, i.e. independent of $\theta_i$, we can ignore them, which leaves us with Equation \ref{prejoin}:
\begin{equation}
\label{prejoin}
L(\theta_i) = (1+\delta)\mathcal{Q}(s_j,a_j;\theta_i)^2 - 2(\hat{y} + \delta\bar{y})\mathcal{Q}(s_j,a_j;\theta_i)
\end{equation}
In order to complete the squares, we introduce the constant $(\hat{y} + \delta\bar{y})^2$, and divide by $(1+\delta)$ to obtain the final loss as Equation \ref{finalloss}:
\begin{equation}
\label{finalloss}
L(\theta_i) = \bigl(\overbrace{\frac{\hat{y} + \delta\bar{y}}{1+\delta}}^{r_\text{new}} - \mathcal{Q}(s_j,a_j;\theta_i)\bigr)^2
\end{equation}
With this formulation, ${r_\text{new}}$ can be considered as the smoothened reward, a balance between the current discounted reward and the previous action-value.
Notice that the proposed model doesn't introduce any additional computations, as both loss terms $L^D$ and $L$
share the same set of gradients. which results in the following parameter update, Equation \ref{grad}:
\begin{equation}
\label{grad}
\theta_{i+1} = \theta_i + \alpha\bigl(r_\text{new}- \mathcal{Q}(s_j,a_j;\theta_i)\bigr)\nabla_{\theta_i}\mathcal{Q}(s_j,a_j;\theta_i),
\end{equation}
with $\alpha$ as the scalar step size.
\begin{algorithm}
\caption{in Hindsight Algorithm}\label{alg:hindsight}
\begin{algorithmic}[1]  
	\State\textit{Randomly initialize $\mathcal{Q}(s,a;\theta)$ and $\mathcal{Q}_{target}(s,a;\theta^-)$}
	\State\textit{Initialize modified experience replay buffer $\mathcal{B}$}	
	\For{\textit{episodes $e = 1,M$}}
		\State\textit{Initialize environment $\mathcal{E}$}
		\For{t$=1,T$}
			\State Determine and Execute action $a_t$
			\begin{equation*}
			a_t = 
			\begin{cases}
					\max_a \mathcal{Q}(s_i,a;\theta_i),& \text{if } prob \geq \epsilon\\
					\text{random},              & \text{otherwise}
			\end{cases}			
			\end{equation*}
			\State Receive reward $r_i$ and new state $s_{i+1}$
			\State Store experiences $(s_i,s_{i+1},a_i,\mathcal{Q}(s_{i},a_i;\theta_i),r_i)$ in $\mathcal{B}$
			\State Sample a minibatch of experiences from $\mathcal{B}$
			\State Set target value $\hat{y}$
			\begin{equation*}
			\hat{y} = 
			\begin{cases}
					r_j,& \hspace{-2cm}\text{for terminal state }s_{j+1}\\
					r_j + \gamma \max_a \mathcal{Q}_\text{target}(s_{j+1},a;\theta^-),  & \text{otherwise}
			\end{cases}			
			\end{equation*}
			with $j$ as the index of the sampled observation
			\State Update the network by minimizing
			\begin{equation*}
				L(\theta_i) = (\hat{y} -\mathcal{Q}(s_{j},a_j;\theta_i))^2 +  \delta(\mathcal{Q}(s_{j},a_j;\theta_j) -\mathcal{Q}(s_{j},a_j;\theta_i))^2
			\end{equation*}
			or,
			\begin{equation*}
				L(\theta_i) = (r_\text{new} -\mathcal{Q}(s_{j},a_j;\theta_i))^2
			\end{equation*}			
		\EndFor
	\EndFor
\end{algorithmic}
\end{algorithm}
\\In order to implement this algorithm, we modify the experience replay buffer, $\mathcal{B}$ to accommodate the action-values of the states. Hence at every frame, we store the transitions ($s_j,s_{j+1},a_j,\mathcal{Q}(s_j,a_j;\theta_j),r_j$) in the memory. The goal is to improve the performance of the Q-function, by introducing updates that do not emphasize solely on the future discounted reward, but also take into account not to deviate from the values associated with decisions in the agent's experience in older encounters, and reduces overestimation errors.
\section{Overestimation and Approximation Errors}
One of the issues of function estimation based on Q-learning is the overestimation phenomenon \cite{thrun1993issues} that lead asymptotically to sub-optimal policies. Assuming action-values are corrupted by uniformly distributed noise in an interval $[-\epsilon,\epsilon]$, target values would be overestimated by a value with an upper bound of $\gamma\epsilon\frac{m-1}{m+1}$, due to the max operator, with $\gamma$ as the discount factor and $m$ as the number of actions. Overestimations also have a tight lower bound \cite{van2016deep}, which is derived as $\sqrt[]{\frac{C}{m-1}}$, with $C > 0$. The DDQN approach reduces overestimation, and replaces the positive bias with a negative one. \\
The effect of the hindsight factor on overestimation is demonstrated in the following function estimation experiment\cite{van2016deep}. The environment is described as a continuous real-valued state-space with 10 discrete actions per state. Each action represents a polynomial function, with a chosen degree of 6, fitted to a subset of integer states, with two adjacent states missing; for action $a_0$, states -5 and -4 are removed, for action $a_1$, states -4 and -3, and so on. Each action has the same true value, defined as either $Q_*(s,a)=sin(s)$ or $Q_*(s,a)=2\exp(-s^2)$.\\
We are able to reproduce the experiment for DQN and DDQN and obtain the exact overestimation values, presented in the original approximation \cite{van2016deep}, as can be seen in Figure \ref{overest}. Systemic overestimation is an artifact of recursive function approximation, which leads to a detorioration of value estimates as the action values are assumed true, when in fact they contain noise. Introducing the hindsight factor maintains low bias in the estimates, especially when applied to DQN. We also notice, that even though the bias is slightly higher than DDQN, it is indeed however much smoother, which translates into an overall better estimation. However, we realize later that applying the hindsight factor to DDQN can in some cases lead to an extremely cautious exploration within the game, and respectively lower rewards. 
\begin{table}
\centering
\caption{Summarized performance for 33 games}
\begin{tabular}{ l | c  r }
Method&DQN&DQN-H\\
Wins w.r.t all&2&4\\
Wins w.r.t counterpart&10&23\\
Score&676&2874\\
\hline
Method&DDQN&DDQN-H\\
Wins w.r.t all&2&2\\
Wins w.r.t counterpart&15&18\\
Score&1632&2593\\
\hline
Method&DUEL&DUEL-H\\
Wins w.r.t all&6&17\\
Wins w.r.t counterpart&9&24\\
Score&3247&4342\\
\end{tabular}
\label{preview}
\end{table}
\begin{figure*}[h]
\subfloat{%
  \includegraphics[width=0.7\columnwidth]{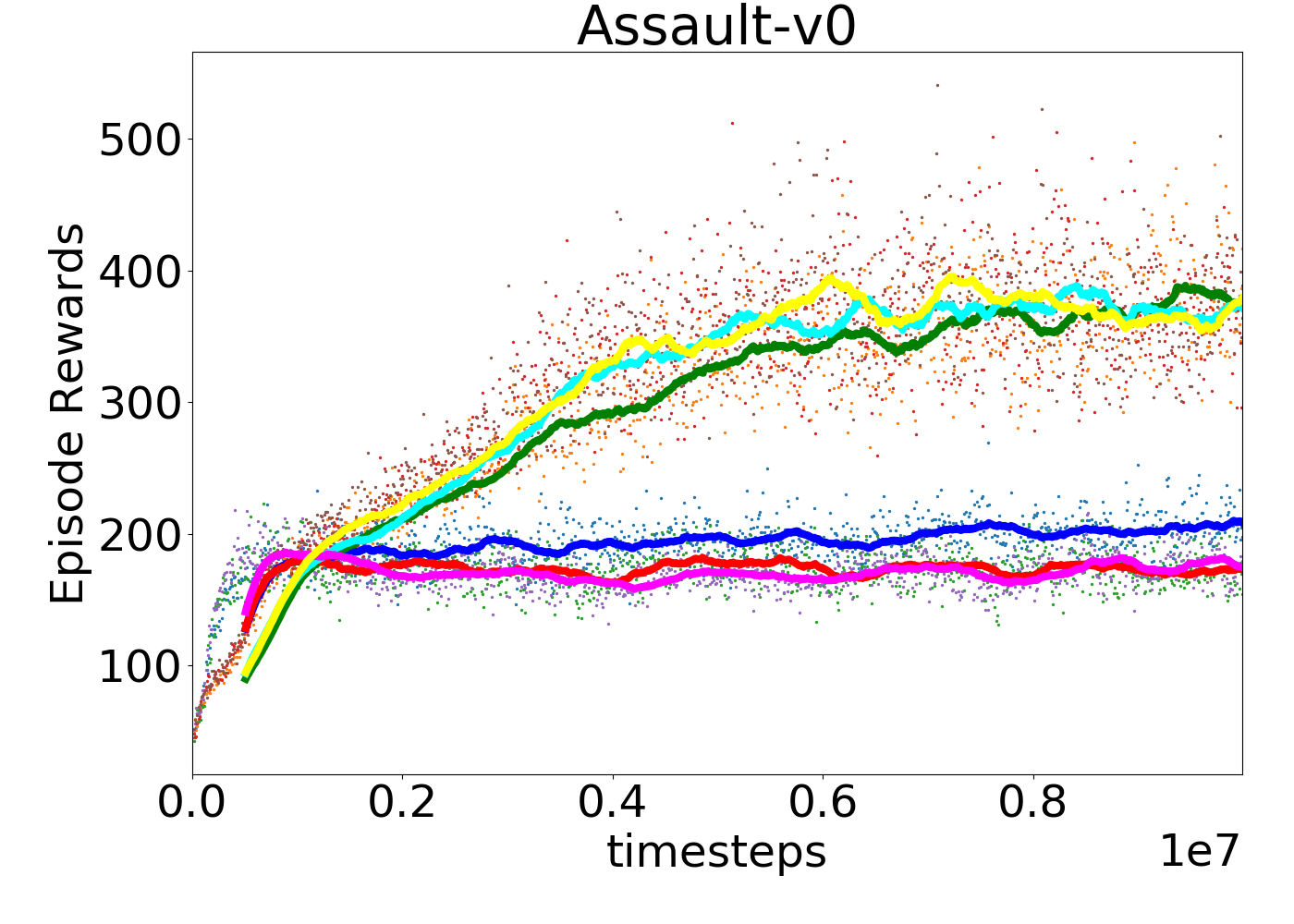}%
}
\subfloat{%
  \includegraphics[width=0.7\columnwidth]{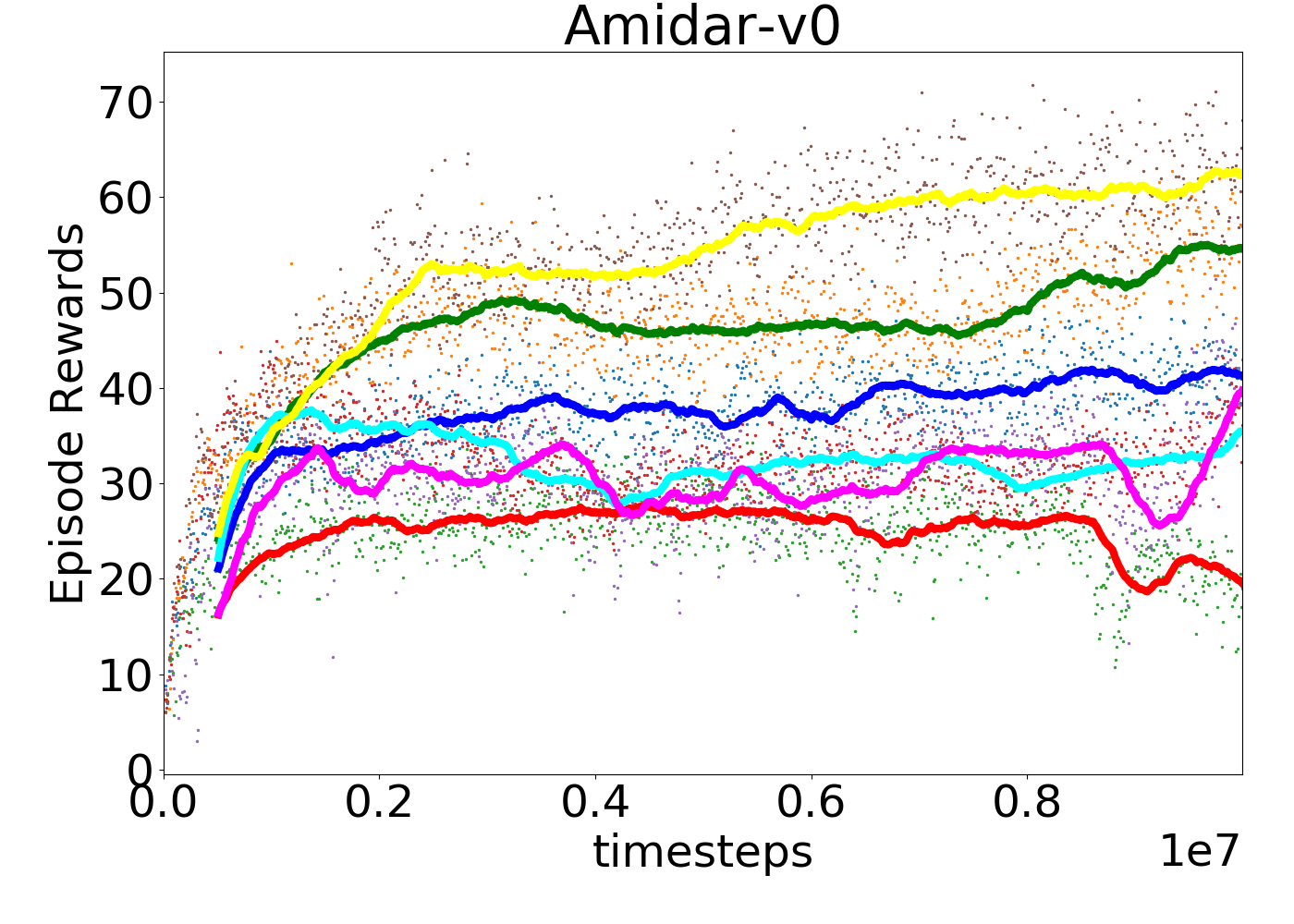}%
}
\subfloat{%
  \includegraphics[width=0.7\columnwidth]{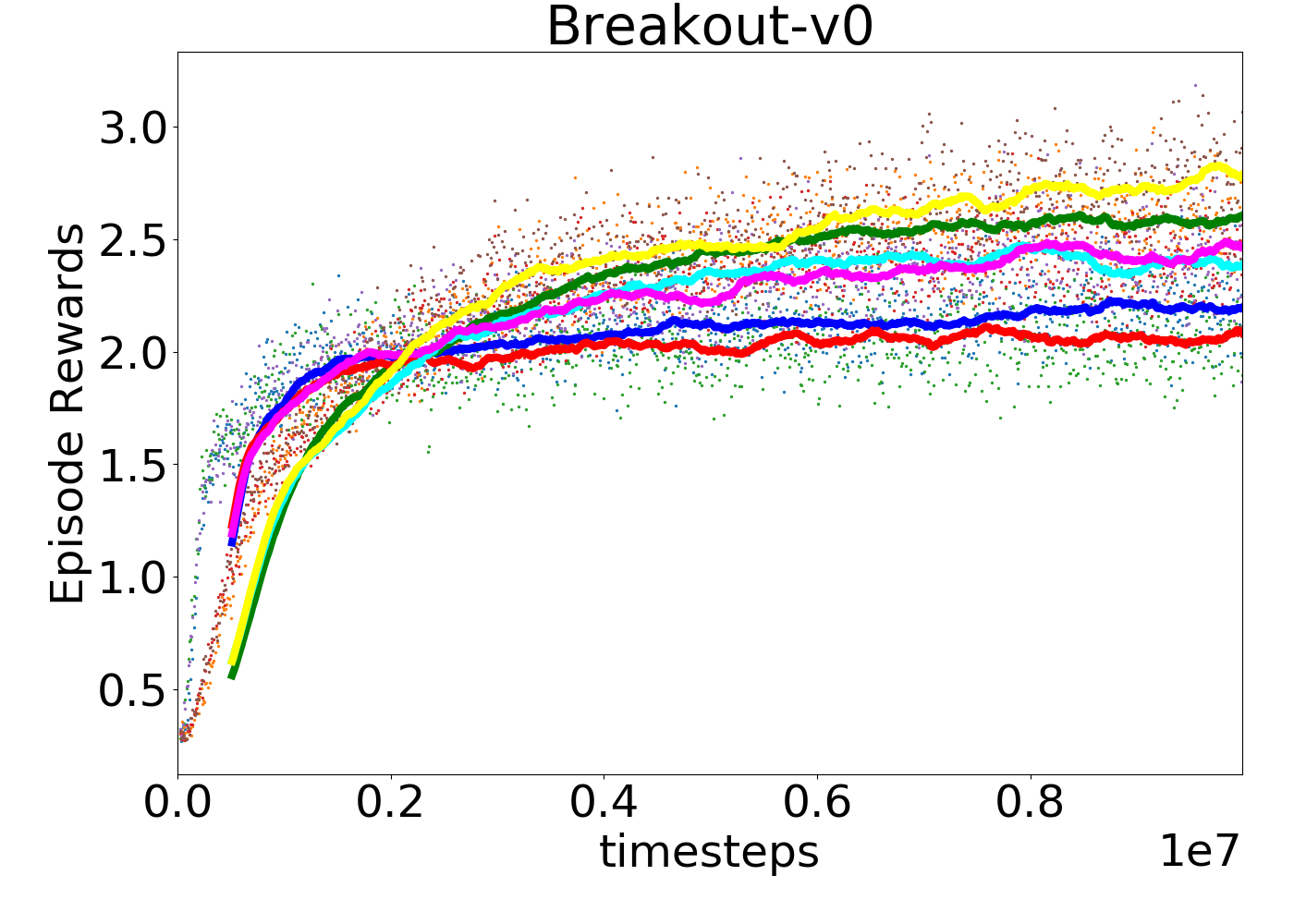}%
}
\\
\subfloat{%
  \includegraphics[width=0.7\columnwidth]{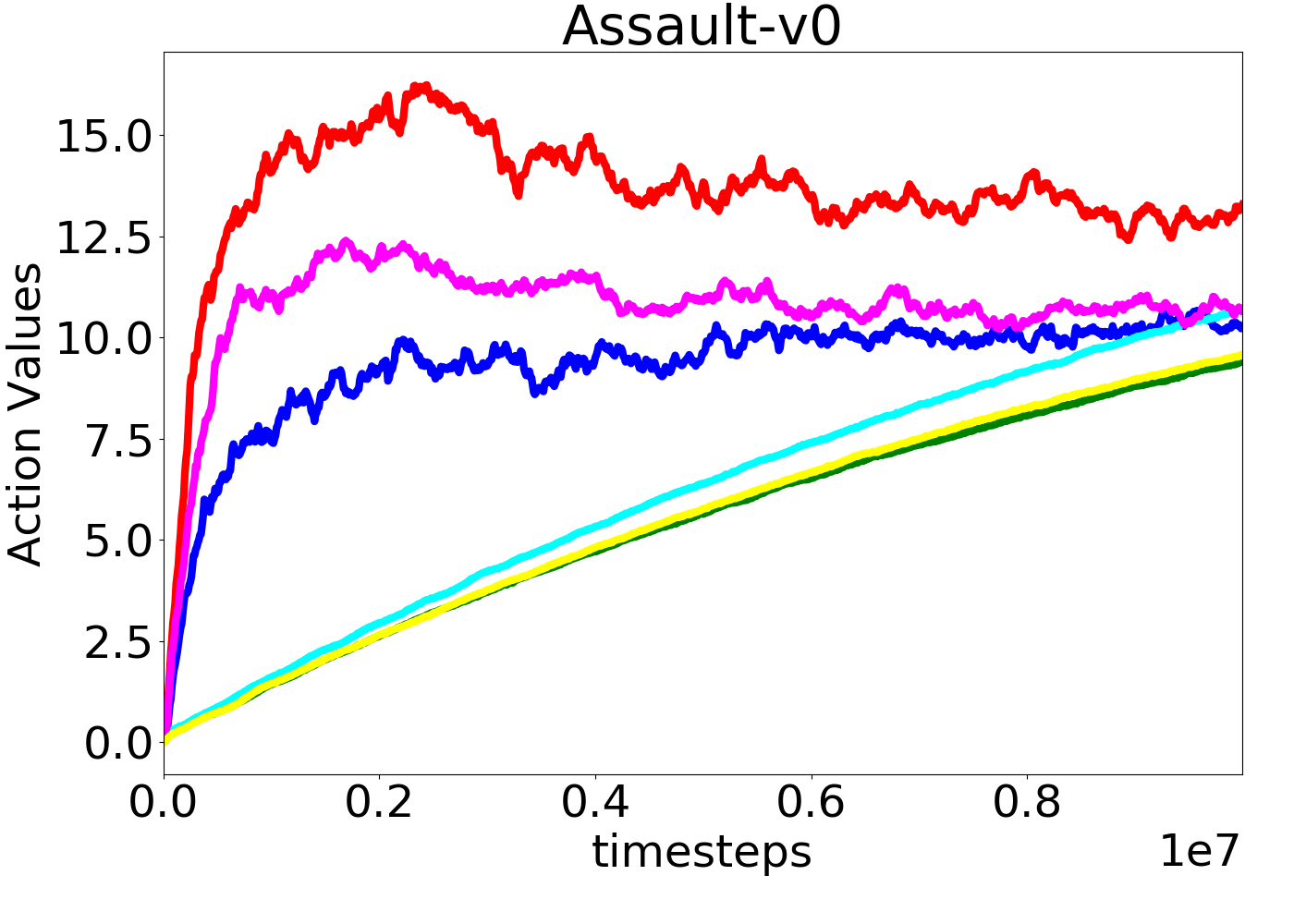}%
}
\subfloat{%
  \includegraphics[width=0.7\columnwidth]{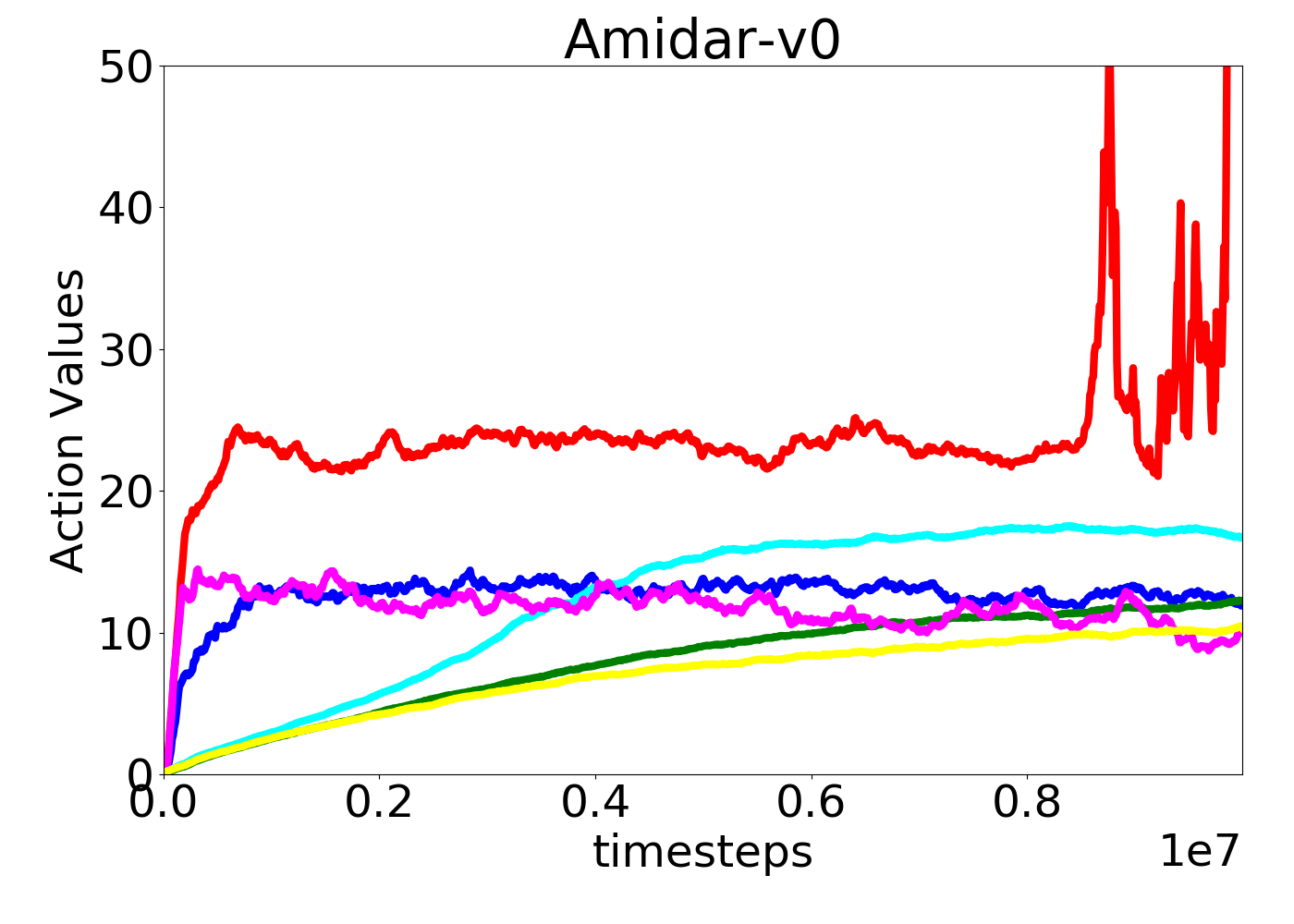}%
}
\subfloat{%
  \includegraphics[width=0.7\columnwidth]{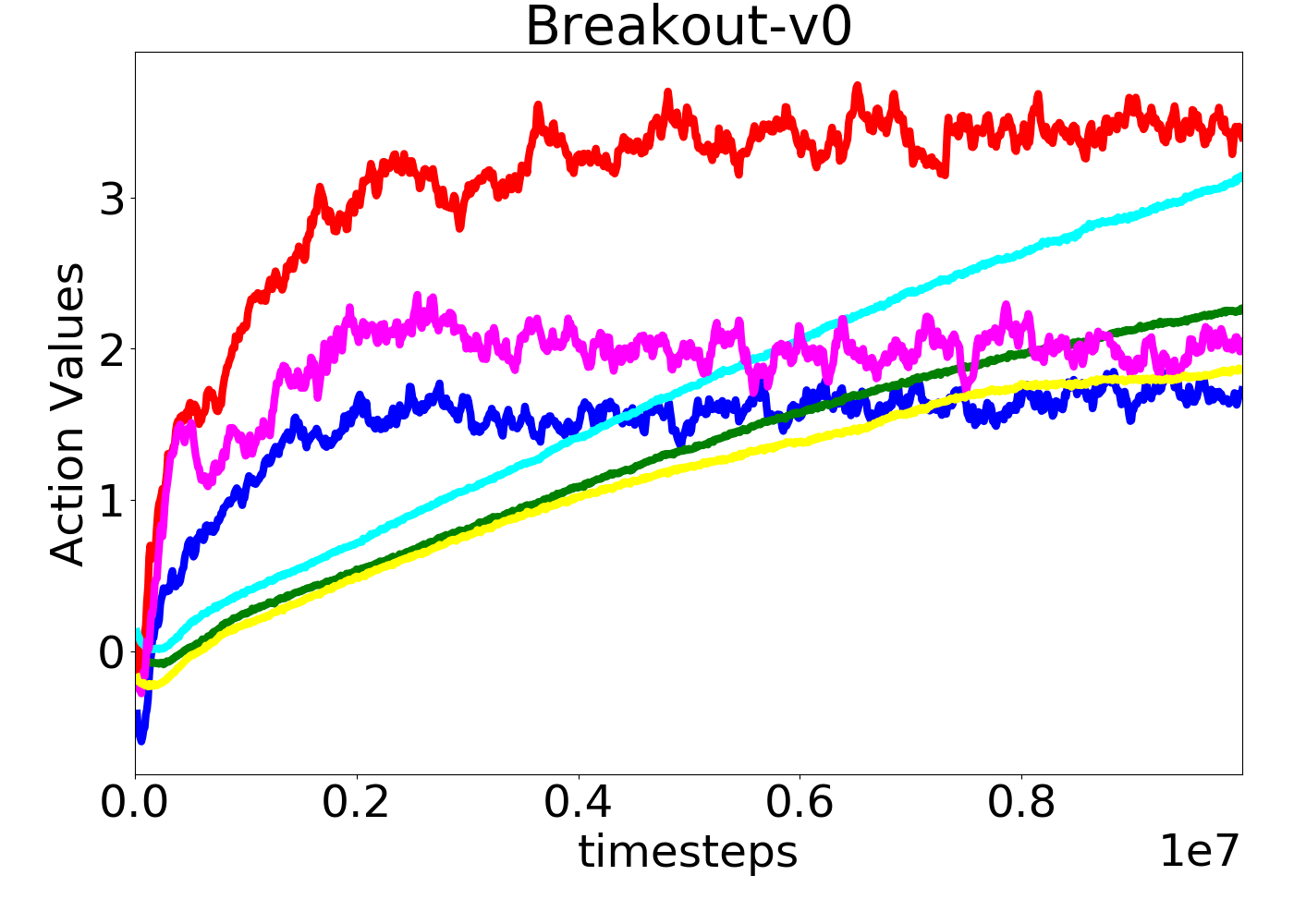}%
}
\caption{Performance curves for various ATARI games using variants of Q-learning techniques; DDQN(dark blue),DDQN-H(green);DQN(red),DQN-H(cyan); DUEL(pink),DUEL-H(yellow)}
\label{perf}
\end{figure*}
\section{Experimental Results}
\label{experiments}
We now demonstrate the practical advantage of adding the hindsight factor to the Q-learning loss function. To do so, we reimplement several variations of deep Q-learning methods, namely: DQN \cite{mnih2015human}, Double DQN \cite{van2016deep}, and dueling networks \cite{wang2015dueling}. All the models are trained in TensorFlow \cite{abadi2016tensorflow} on a GeForce 1080 TI GPU, using the hyperparameters provided by \cite{mnih2015human}, with the average runtime duration per baseline amounting to 30 GPU hours. The proposed model modifies these existing architectures by introducing the hindsight factor as an additional loss term, and is referred to as $\langle BASE \rangle$-H. As shown in Algrorithm \ref{alg:hindsight}, the buffer is extended to accommodate the action-value per state. We evaluated the proposed method on more than 30 ATARI games, which differ in terms of difficulty, number of actions, as well as the importance of memory, i.e. previous state-action values. We do not report on the games that did not achieve any significant learning for the specified number of frames. The results showcase the importance of the hindsight factor under various settings, and its contribution to an overall improved performance over the deep Q-network counterparts. Table \ref{preview} summarizes the mean score of 100 episodes after training for 10 million frames.

Figure \ref{perf} represents the performance curves of the baselines and the proposed approach for a selection of games. 
The hindsight factor has a different effect on every approach depending on the nature of the game. However, the results are clearly indicative of the power of hindsight. Conventional Q-learning techniques lead to an early rise in performance, which is attributed to a more courageous exploration, as compared to a delayed increase in the reward when using the hindsight factor, attributed to the cautious exploration. However, as the learning algorithm progresses, the baselines, seem to plateau at a local optima, as the performance remains consistent for several million frames.\\
As mentioned earlier, the hindsight factor models an adaptive learning rate controller, further discussed in the following section. This is also  realized experimentally, for example in AMIDAR, when DQN performance detoriorates (in red) at the final one million frames, whereas DQN-H keeps on improving which is also a sign that introducing the hindsight factor prevents overfitting. Even for simple games such as BREAKOUT, the relative difference in performance between the proposed approach and the counterpart baseline is significant.
\\
We also take a look at the values of the selected actions using the hindsight method. Smoothing the discounted reward by previous reward values turns out to have a great impact on the action values. Throughout the training process, the action-values selected by applying the hindsight factor seem to increase at a steady (linear) pace with no signs of convergance as the number of frames exceeds ten million frames. The opposite can be said about the regular Q-learning techniques. Overestimation in standard Q-learning can be avoided with DDQN, where we notice that it results in the lower set of action-values as compared to DQN and DUELING. Nevertheless, these values are still higher than their hindsight counterparts, especially at the early stages of learning, which proves that there are still some reminent overestimation inherent in DDQN. 
\subsubsection{The underlying effect}
\label{udefect}
\begin{figure}[h]
\centering
\subfloat{%
  \includegraphics[width=0.7\columnwidth]{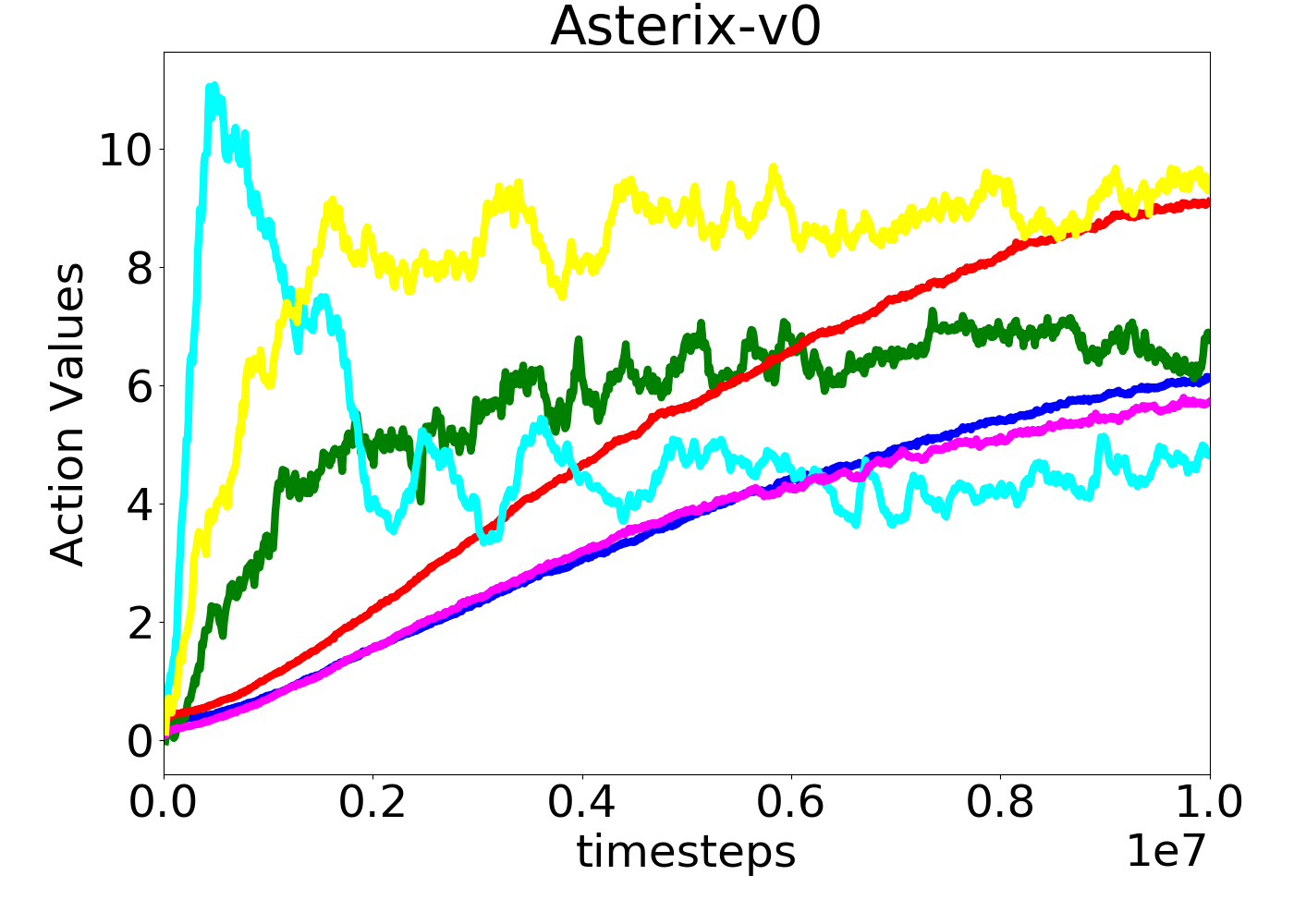}%
}
\caption{Performance curves for ASTERIX where the baselines have lower learning rate. DQN-H(cyan), DDQN-H(green), DUEL-H(yellow), DQN-H-HALF(red), DDQN-H-HALF(dark blue), DUEL-H-HALF(pink)}
\label{diflr}
\end{figure}

The underlying effect of the hindsight factor is that it adaptively changes the learning rate, as it establishes a direct dependance on the action value, unlike exsiting adaptive optimizers such as ADAM \cite{kingma2014adam} and RMSProp \cite{rmsprop}, which depend only on the evolution of the gradients. To estimate the value of state-action pairs in a discounted Markov Decision Processes, Equation \ref{watlr} is introduced \cite{watkins1992q}:
\begin{equation}
\label{watlr}
\begin{split}
Q_{i+1}(s,a) = & (1-\alpha_{i}(s,a))Q_{i}(s,a) + \\ &\alpha_i(s,a)\underbrace{(r_i(s,a)+\gamma \max_{b\in A}Q_i(s',b))}_\text{Target Reward},
\end{split}
\end{equation}
where $\alpha_i$ represents the learning rate at iteration $i$ and $s'$ represents the next state resulting from action $a$ at state $s$. Given the hindsight factor, we replace the target reward with $r_\text{new}$, which leads to,
\begin{equation}
\label{watlrupdated}
\begin{split}
Q_{i+1}(s&,a)=\bigl(1-\alpha_{i}(s,a)\bigr)Q_{i}(s,a) +\\&\frac{\alpha_i(s,a)}{1+\delta}\bigl(r_i(s,a)+\gamma \max_{b \in A}Q_i(s',b) + \delta Q_{j}(s,a))\bigr)
\end{split}
\end{equation}
with $j<i$.
%After reordering the variables, the state-action update becomes,
%\begin{equation}
%\label{watlrnew}
%\begin{split}
%Q_{i+1}&(s,a) =\bigl(1-\frac{\alpha_{i}(s,a)}{1+\delta}\bigr)Q_{i}(s,a) + \\&\frac{\alpha_i(s,a)}{1+\delta}\bigl(r_i(s,a)+\gamma \max_{b\in A}Q_i(s',b)\bigr).
%\end{split}
%\end{equation}
The effect of introducing $Q_{j}(s,a)$ is an adaptive state-action pair updates that dynamically changes over time. Now if we replace $Q_{j}(s,a)$ with $Q_{i}(s,a)$, i.e. the effect would be scaling down the learning rate by a factor of $\frac{1}{1+\delta}$. Nevertheless, the learning rate would still be fixed and does not adapt to the change in model parameters. We highlight this effect in Figure \ref{diflr} as we see that with a halved learning rate results in a better performance for DQN and DEUL, however still results in higher overestimation errors and plateaus at an early stage. Reducing the learning rate partially models the effect provided by the additonal loss, as it is still independent of the action-values, and does not particularly help with overfitting. This is evident by the early spike of the DQN at during the first two million frames of training.

\subsubsection{Adjusting the Hindsight coefficient}
\begin{figure}[h]
\subfloat{%
  \includegraphics[width=0.5\columnwidth]{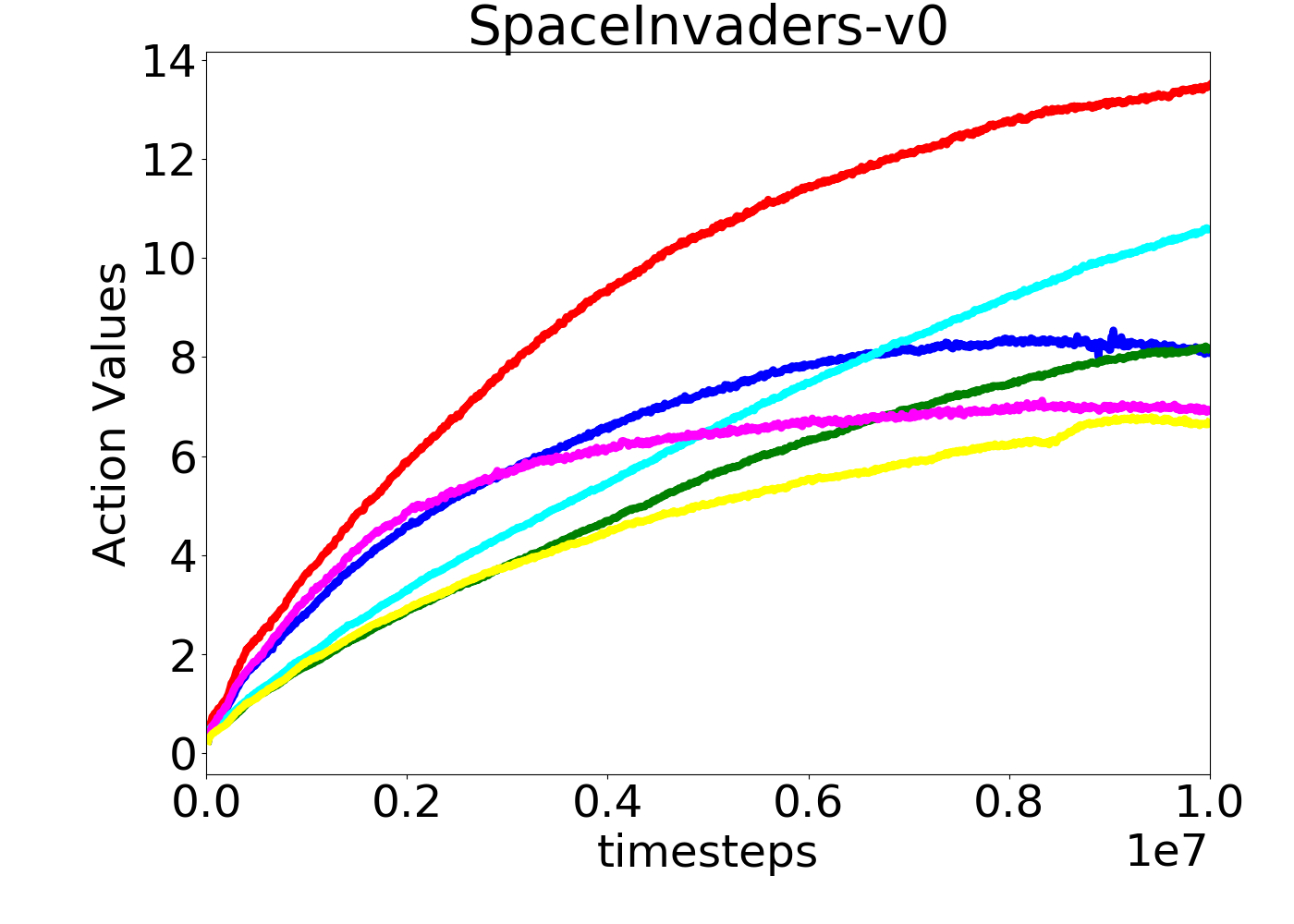}%
}
\subfloat{%
  \includegraphics[width=0.5\columnwidth]{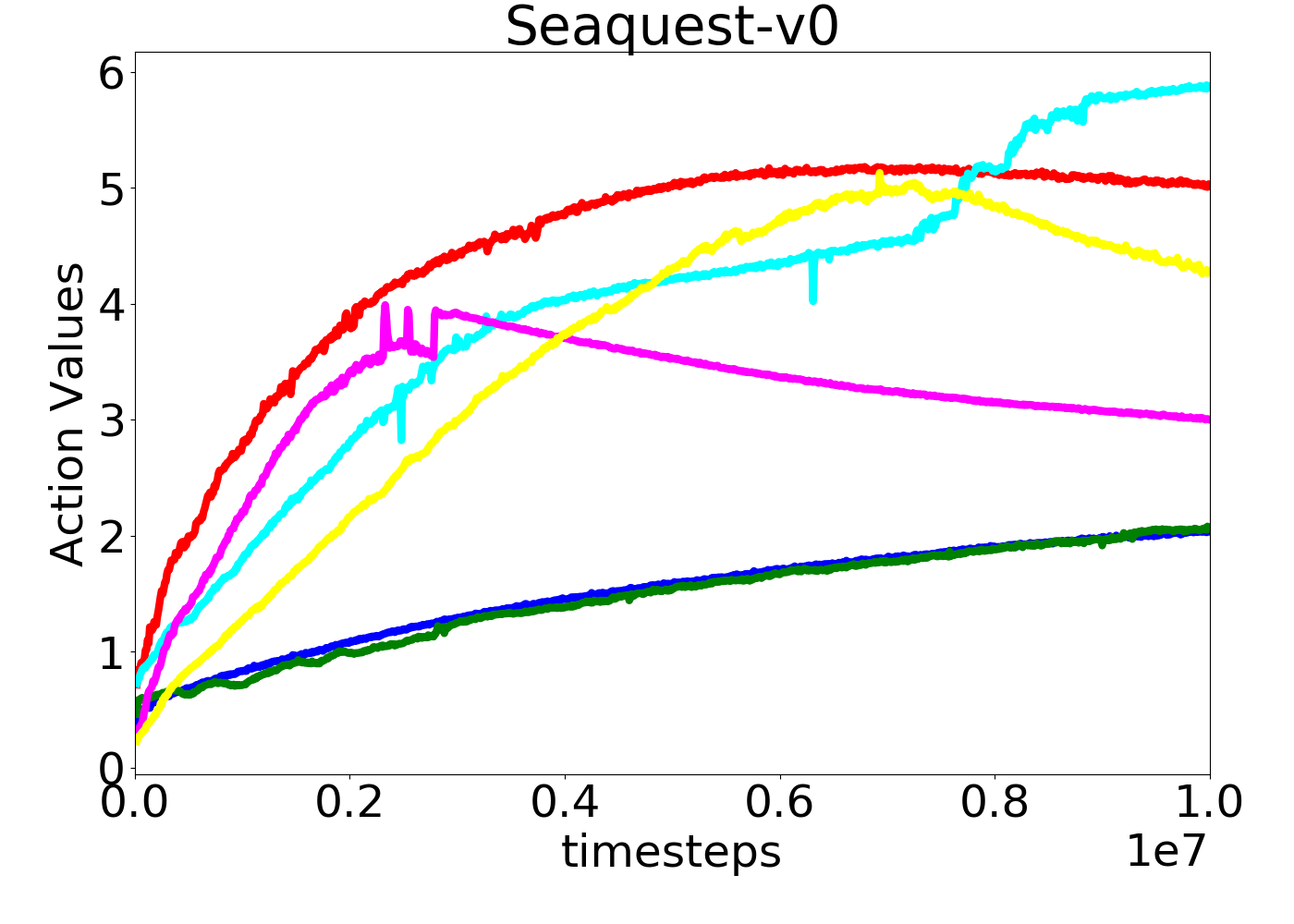}%
}
\caption{Performance curves with $\delta=1$ and $\delta=\frac{1}{2}$; DQN-H(cyan), DDQN-H(green), DUEL-H(yellow), DQN-H-HALF(red), DDQN-H-HALF(dark blue), DUEL-H-HALF(pink)}
\label{half}
\end{figure}

For the previous experiments, we have fixed the hindsight coefficient $\delta$ to 1, and hence uniformly weighing the reward between the expected gain and historic achievement. In the following, we juxtapose the performance obtained by setting $\delta=\frac{1}{2}$, and $\delta=-\frac{1}{2}$.
First we notice that if we set the hindsight coefficent to $-\frac{1}{2}$, the agent is prone to diverge almost immediately, so no results are shown. On the other hand, setting the hindsight coefficient to $\frac{1}{2}$, as expected, results in slightly higher action values that is caused by the decreased dependence of new action-values on the history, and hence allowing for more overestimation, Figure \ref{half}. A lower hindsight coefficient has a positive impact at the early stages of learning when the agent is still exploring the environment, however, as the number of frames increases, the agent becomes prone to overfitting, and ultimately results in a lower performance.
\\Optimizing the Q-function using the hindsight factor as a regularizer to smoothen the expected reward turns out to improve the performance well before the action-values seem to converge. However, with some games we notice that the performance is negatively effected by this formulation. This might be attributed to the penalty which the hindsight factor indirectly applies on exploration. In addition, it is worth noting that as the hindsight coefficient decreases to $0$, the action-values start to come closer to their counterpart models.
\section{Conclusion}
Existing Q-learning techniques aim at maximizing the expected reward, by minimizing the difference between the current action-value and the expected discounted reward. 
However, they offer no insight into the past as the progress of the estimator, measured through the difference between the current action-value and the action-value at the same state at a previous iteration, is ignored. 
In this paper, we proposed the introduction of the hindsight factor, an additional loss function that shares the same gradients of the prediction network, and hence incurring no extra computational efforts. The hindsight factor acts a reward regularizer, forcing the reward to be more realistic and hence avoiding overestimation. The new reward is a trade-off between the expected discounted reward, and the historic temporal difference. Through a deterministic function estimation problem, we are able to prove that by adding the hindsight factor to exiting function estimators via Q-learning, we are able to reduce the average error, and produce a stable estimation. The underlying effect of the hindsight factor is translated as an adaptively controlled learning rate that outperforms the respective base models. We have shown that in general outperforms deep Q-networks, double deep Q-networks and dueling networks in $74\%$, $58\%$, and $77\%$  of the games, respectively. \\Moving forward, it would be interesting to study the effect of introducing an adaptive hindsight coefficient, based on the absolute reward improvement across frames.
\clearpage
\bibliographystyle{named}
\bibliography{jomaa2019a-ijcai}

\begin{thebibliography}{}

\bibitem[\protect\citeauthoryear{Abadi \bgroup \em et al.\egroup
  }{2016}]{abadi2016tensorflow}
Mart{\'\i}n Abadi, Paul Barham, Jianmin Chen, Zhifeng Chen, Andy Davis, Jeffrey
  Dean, Matthieu Devin, Sanjay Ghemawat, Geoffrey Irving, Michael Isard, et~al.
\newblock Tensorflow: a system for large-scale machine learning.
\newblock In {\em OSDI}, volume~16, pages 265--283, 2016.

\bibitem[\protect\citeauthoryear{Anschel \bgroup \em et al.\egroup
  }{2017}]{anschel2017averaged}
Oron Anschel, Nir Baram, and Nahum Shimkin.
\newblock Averaged-dqn: Variance reduction and stabilization for deep
  reinforcement learning.
\newblock In {\em Proceedings of the 34th International Conference on Machine
  Learning-Volume 70}, pages 176--185. JMLR. org, 2017.

\bibitem[\protect\citeauthoryear{Bellman}{1957}]{bellman1957functional}
Richard Bellman.
\newblock Functional equations in the theory of dynamic programming--vii. a
  partial differential equation for the fredholm resolvent.
\newblock {\em Proceedings of the American Mathematical Society},
  8(3):435--440, 1957.

\bibitem[\protect\citeauthoryear{Farebrother \bgroup \em et al.\egroup
  }{2018}]{farebrother2018generalization}
Jesse Farebrother, Marlos~C Machado, and Michael Bowling.
\newblock Generalization and regularization in dqn.
\newblock {\em arXiv preprint arXiv:1810.00123}, 2018.

\bibitem[\protect\citeauthoryear{Hinton \bgroup \em et al.\egroup
  }{2012}]{rmsprop}
Geoffrey Hinton, N~Srivastava, and Kevin Swersky.
\newblock Lecture 6a overview of mini–batch gradient descent.
\newblock \url{https://class.coursera.org/neuralnets-2012-001/lecture}, 2012.
\newblock Online.

\bibitem[\protect\citeauthoryear{Kingma and Ba}{2014}]{kingma2014adam}
Diederik~P Kingma and Jimmy Ba.
\newblock Adam: A method for stochastic optimization.
\newblock {\em arXiv preprint arXiv:1412.6980}, 2014.

\bibitem[\protect\citeauthoryear{Lin}{1992}]{lin1992self}
Long-Ji Lin.
\newblock Self-improving reactive agents based on reinforcement learning,
  planning and teaching.
\newblock {\em Machine learning}, 8(3-4):293--321, 1992.

\bibitem[\protect\citeauthoryear{Lin}{1993}]{lin1993reinforcement}
Long-Ji Lin.
\newblock Reinforcement learning for robots using neural networks.
\newblock Technical report, Carnegie-Mellon Univ Pittsburgh PA School of
  Computer Science, 1993.

\bibitem[\protect\citeauthoryear{Mnih \bgroup \em et al.\egroup
  }{2015}]{mnih2015human}
Volodymyr Mnih, Koray Kavukcuoglu, David Silver, Andrei~A Rusu, Joel Veness,
  Marc~G Bellemare, Alex Graves, Martin Riedmiller, Andreas~K Fidjeland, Georg
  Ostrovski, et~al.
\newblock Human-level control through deep reinforcement learning.
\newblock {\em Nature}, 518(7540):529, 2015.

\bibitem[\protect\citeauthoryear{Schaul \bgroup \em et al.\egroup
  }{2015}]{schaul2015prioritized}
Tom Schaul, John Quan, Ioannis Antonoglou, and David Silver.
\newblock Prioritized experience replay.
\newblock {\em arXiv preprint arXiv:1511.05952}, 2015.

\bibitem[\protect\citeauthoryear{Schmidhuber}{1991}]{schmidhuber1991curious}
J{\"u}rgen Schmidhuber.
\newblock Curious model-building control systems.
\newblock In {\em Neural Networks, 1991. 1991 IEEE International Joint
  Conference on}, pages 1458--1463. IEEE, 1991.

\bibitem[\protect\citeauthoryear{Sutton and
  Barto}{2018}]{sutton2018reinforcement}
Richard~S Sutton and Andrew~G Barto.
\newblock {\em Reinforcement learning: An introduction}.
\newblock MIT press, 2018.

\bibitem[\protect\citeauthoryear{Thrun and Schwartz}{1993}]{thrun1993issues}
Sebastian Thrun and Anton Schwartz.
\newblock Issues in using function approximation for reinforcement learning.
\newblock In {\em Proceedings of the 1993 Connectionist Models Summer School
  Hillsdale, NJ. Lawrence Erlbaum}, 1993.

\bibitem[\protect\citeauthoryear{Van~Hasselt \bgroup \em et al.\egroup
  }{2016}]{van2016deep}
Hado Van~Hasselt, Arthur Guez, and David Silver.
\newblock Deep reinforcement learning with double q-learning.
\newblock In {\em AAAI}, volume~2, page~5. Phoenix, AZ, 2016.

\bibitem[\protect\citeauthoryear{Wang \bgroup \em et al.\egroup
  }{2015}]{wang2015dueling}
Ziyu Wang, Tom Schaul, Matteo Hessel, Hado Van~Hasselt, Marc Lanctot, and Nando
  De~Freitas.
\newblock Dueling network architectures for deep reinforcement learning.
\newblock {\em arXiv preprint arXiv:1511.06581}, 2015.

\bibitem[\protect\citeauthoryear{Watkins and Dayan}{1992}]{watkins1992q}
Christopher~JCH Watkins and Peter Dayan.
\newblock Q-learning.
\newblock {\em Machine learning}, 8(3-4):279--292, 1992.

\end{thebibliography}

\end{document}